\documentclass[11pt,a4paper]{article}
\usepackage[hyperref]{acl2018}
\usepackage{times}
\usepackage{latexsym}
\usepackage{amsmath,graphicx,lipsum,xr}
\usepackage{diagbox}
\usepackage{placeins}
\usepackage{url}

\aclfinalcopy 

\title{Neural User Simulation for Corpus-based Policy Optimisation for \\ Spoken Dialogue Systems}
  
\author{\large Florian L. Kreyssig, I{\~n}igo Casanueva\\  {\bf Pawe{\l} Budzianowski \and Milica Ga{\v{s}}i{\'c}}\\
  Cambridge University Engineering Department,
\\
  Trumpington Street, Cambridge, CB2 1PZ, UK \\
  {\tt \{flk24,ic340,pfb30,mg436\}@cam.ac.uk} \\}

\date{}

\begin{document}
\maketitle
\begin{abstract}
User Simulators are one of the major tools that enable offline training of task-oriented dialogue systems. For this task the Agenda-Based User Simulator (ABUS) is often used. The ABUS is based on hand-crafted rules and its output is in semantic form. Issues arise from both properties such as limited diversity and the inability to interface a text-level belief tracker. This paper introduces the Neural User Simulator (NUS) whose behaviour is learned from a corpus and which generates natural language, hence needing a less labelled dataset than simulators generating a semantic output. In comparison to much of the past work on this topic, which evaluates user simulators on corpus-based metrics, we use the NUS to train the policy of a reinforcement learning based Spoken Dialogue System. The NUS is compared to the ABUS by evaluating the policies that were trained using the simulators. Cross-model evaluation is performed i.e. training on one simulator and testing on the other. Furthermore, the trained policies are tested on real users. In both evaluation tasks the NUS outperformed the ABUS.
\end{abstract}

\section{Introduction}
Spoken Dialogue Systems (SDS) allow human-computer interaction using natural speech. Task-oriented dialogue systems, the focus of this work, help users achieve goals such as finding restaurants or booking flights \cite{young2013pomdp}.

Teaching a system how to respond appropriately in a task-oriented setting is non-trivial. In state-of-the-art systems this \textit{dialogue management} task is often formulated as a reinforcement learning (RL) problem \cite{young2013pomdp, roy2000spoken, williams2007partially,gasic2014gaussian}. In this framework, the system learns by a \textit{trial and error} process governed by a 
\textit{reward function}. User Simulators can be used to train the policy of a \textit{dialogue manager} (DM) without 
real user interactions. Furthermore, they allow an unlimited number of dialogues to be created with each dialogue being faster than a dialogue with a human.

In this paper the Neural User Simulator (NUS) is introduced which outputs natural language and whose behaviour is learned from a corpus. 
The main component, inspired by \cite{el2016sequence}, consists of a feature extractor and a neural network based sequence-to-sequence model \cite{sutskever2014sequence}. The sequence-to-sequence model consists of a recurrent neural network (RNN) encoder that encodes the dialogue history and a decoder RNN which outputs natural language. Furthermore, the NUS generates its own goal and possibly changes it during a dialogue. This allows the model to be deployed for training more sophisticated DM policies. To achieve this, a method is proposed that transforms the goal-labels of the used dataset (DSTC2) into labels whose behaviour can be replicated during deployment.

The NUS is trained on dialogues between real users and an SDS in a restaurant recommendation domain. Compared to much of the related work on user simulation, we use the trained NUS to train the policy of a reinforcement learning based SDS. In order to evaluate the NUS, an Agenda-Based User-Simulator (ABUS) \cite{schatzmann2007agenda} is used to train another policy. The two policies are compared against each other by using \textit{cross-model} evaluation \cite{schatztmann2005effects}. This means to train on one model and to test on the other. Furthermore, both trained policies are tested on real users. On both evaluation tasks the NUS outperforms the ABUS, which is currently one of the most popular off-line training tools for reinforcement learning based Spoken Dialogue Systems \cite{koo2015implementation,fatemi2016policy,chen2017agent,chang2017affordable,Casanueva2018feudal,weisz2018sample,shah2018building}.

The remainder of this paper is organised as follows. Section 2 briefly describes task-oriented dialogue. Section 3 describes the motivation for the NUS and discusses related work. Section 4 explains the structure of the NUS, how it is trained and how it is deployed for training a DM's policy. Sections 5 and 6 present the experimental setup and results. Finally, Section 7 gives conclusions.

\section{Task-Oriented Dialogue}

A Task-Oriented SDS is typically designed according to a structured \textit{ontology}, which defines what the system can talk about. In a system recommending restaurants the ontology defines those attributes of a restaurant that the user can choose, called \textit{informable slots} (e.g. different food types, areas and price ranges), the attributes that the user can request, called \textit{requestable slots} (e.g. phone number or address) and the restaurants that it has data about. An attribute is referred to as a \textit{slot} and has a corresponding \textit{value}. Together these are referred to as a \textit{slot-value pair} (e.g. \texttt{area}=north).

Using RL the DM is trained to act such that is maximises the cumulative future reward. The process by which the DM chooses its next action is called its \textit{policy}. A typical approach to defining the reward function for a task-oriented SDS is to apply a small per-turn penalty to encourage short dialogues and to give a large positive reward at the end of each successful interaction.

\section{Motivation and Related Work}
Ideally the DM's policy would be trained by interacting with real users. 
Although there are models that support on-line learning~\cite{gjtk11}, for the majority of RL algorithms, which require a lot of interactions, this is impractical. Furthermore, a set of users needs to be recruited every time a policy is trained. This makes common practices such as hyper-parameter optimization prohibitively expensive. Thus, it is natural to try to learn from a dataset which needs to be recorded only once, but can be used over and over again.

A problem with learning directly from recorded dialogue corpora is that the state space that was visited during the collection of the data is limited; the size of the recorded corpus usually falls short of the requirements for training a statistical DM. However, even if the size of the corpus is large enough the optimal dialogue strategy is likely not to be contained within it.

A solution is to transform the static corpus into a dynamic tool: a \textit{user simulator}. The user simulator (US) is trained on a dialogue corpus to learn what responses a real user would provide in a given dialogue context. The US is trained using supervised learning since the aim is for it to learn \textit{typical} user behaviour. For the DM, however, we want \textit{optimal} behaviour which is why supervised learning cannot be used. By interacting with the SDS, the trained US can be used to train the DM's policy. The DM's policy is optimised using the feedback given by either the user simulator or a separate evaluator. Any number of dialogues can be generated using the US and dialogue strategies that are not in the recorded corpus can be explored. 

Most user-simulators work on the level of user semantics. These usually consist of a \textit{user dialogue act} (e.g. inform, or request) and a corresponding slot-value pair. The first statistical user simulator \cite{eckert1997user} used a simple bi-gram model $P\!\left(a_u \!\mid\! a_m\right)$ to predict the next user act $a_u$ given the last system act $a_m$. It has the advantage of being purely probabilistic and 
domain-independent. However, it does not take the full dialogue history into account and is not conditioned on a goal, leading to incoherent user behaviour throughout a dialogue. \citet{scheffler2000probabilistic,scheffler2001corpus} attempted to overcome goal inconsistency by proposing a graph-based model. However, developing the graph structure requires extensive domain-specific
 knowledge. \citet{pietquin2006probabilistic} combined features from Sheffler and Young's work with Eckert's Model, by conditioning a set of probabilities on an explicit representation of the user goal and memory. 
A Markov Model is also used by \citet{georgila2005learning}. It uses a large feature vector to describe the user's current state, which helps to compensate for the Markov assumption. However, the model is not conditioned on any goal. Therefore, it is not used to train a dialogue policy since it is impossible to determine whether the user goal was fulfilled. A hidden Markov model was proposed by \citet{cuayahuitl2005human}, which was also not used to train a policy. \citet{chandramohan2011user} cast user simulation as an inverse reinforcement learning problem where the user is modelled as a decision-making agent. The 
model did not incorporate a user goal and was hence not used to train a policy. The most prominent user model for policy optimisation is the Agenda-Based User Simulator \cite{schatzmann2007agenda}, which represents the user state elegantly as a stack of necessary user actions, called the \textit{agenda}. The mechanism that generates the user response and updates the agenda does not require any data, though it can be improved using data. The model is conditioned on a goal for which it has update rules in case the dialogue system expresses that it cannot fulfil the goal. \citet{el2016sequence} modelled user simulation as a sequence-to-sequence task. The model can keep track of the dialogue history and user behaviour is learned entirely from data. However, goal changes were not modelled, 
even though a large proportion of dialogues within their dataset (DSTC2) contains goal changes. Their model outperformed the ABUS on statistical metrics, which is not surprising given that it was trained by optimising a statistical metric and the ABUS was not.

The aforementioned work focuses on user simulation at the semantic level. Multiple issues arise from this approach. Firstly, annotating the user-response with the correct semantics is costly. More data could be collected, if the US were to output natural language. Secondly, research suggests that the two 
modules of an SDS performing Spoken Language Understanding (SLU) and belief tracking should be jointly trained as a single entity \cite{mrkvsic2017neural,sun2016recurrent,sun2014sjtu,zilka2015incremental,ramadan2018large}. In fact in the second Dialogue State Tracking Challenge (DSTC2) \cite{henderson2014second}, the data of which this work uses, systems which used no external SLU module outperformed all systems that only used an external SLU Module\footnote{The best-performing models used both.}. Training the policy of a DM in a simulated environment, when also using a joint system for SLU and belief tracking is \textit{not possible} without a US that produces natural language. Thirdly, a US is sometimes augmented with an error model which generates a set of competing hypotheses with associated confidence scores trying to replicate the errors of the speech recogniser. When the error model matches the characteristics of the speech recogniser more accurately, the SDS performs better \cite{williams2008evaluating}. However, speech recognition errors are badly modelled based on user semantics since they arise (mostly) due to the phonetics of the spoken words and not their semantics \cite{goldwater2010words}. Thus, an SDS that is trained with a natural language based error model is likely to outperform one trained with a semantic error model when tested on real users. Sequence-to-sequence learning for word-level user simulation is performed in \cite{crook2017sequence}, though the model is not conditioned on any goal and hence not used for policy optimisation. A word-level user simulator was also used in \cite{li2017end} where it was built by augmenting the ABUS with a natural language generator.

\section{Neural User Simulator}
\begin{figure}[htb!]
  \centering
  \centerline{\includegraphics[width = 0.90\linewidth]{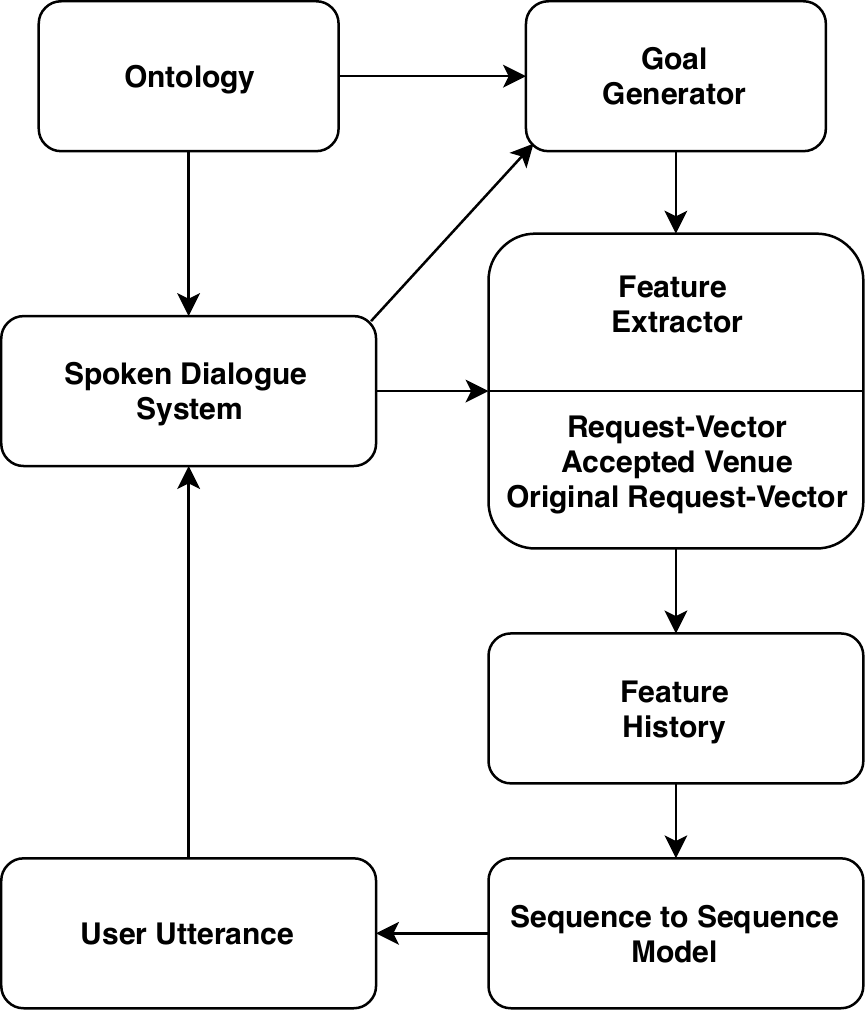}}
  \caption{General Architecture of the Neural User Simulator. The System Output is passed to the Feature Extractor. It generates a new feature vector that is appended to the Feature History, which is passed to the sequence-to-sequence model to produce the user utterance. At the start of the dialogue the Goal Generator generates a goal, which might change during the course of the dialogue.}
\label{fig:NUSfull}
\end{figure}
An overview of the NUS is given in Figure~\ref{fig:NUSfull}. At the start of a dialogue a random goal $G_0$ is generated by the \textit{Goal Generator}. The possibilities for $G_0$ are defined by the \textit{ontology}. In dialogue turn $T$, the output of the SDS ($da_T$) is passed to the NUS's \textit{Feature Extractor}, which generates a feature vector $\mathbf{v}_T$ based on $da_T$, the current user goal, $G_T$, and parts of the dialogue history. This vector is appended to the \textit{Feature History} $\mathbf{v}_{1:T}\!=\!\mathbf{v}_1 ... \mathbf{v}_T$. This sequence is passed to the \textit{sequence-to-sequence} model (Fig.~\ref{fig:NUSseq2seq}), which will generate the user's length $n_T$ utterance $\mathbf{u}_T = w_0 ... w_{n_T}$. As in Figure~\ref{fig:NUSseq2seq}, words in $\mathbf{u}_T$ corresponding to a slot are replaced by a slot token; a process called \textit{delexicalisation}. If the SDS expresses to the NUS that there is no venue matching the NUS's constraints, the goal will be altered by the Goal Generator.
\subsection{Goal Generator}
\begin{table*}[htb!]
\begin{center}
\begin{tabular}{|l|l|l|}
\hline \bf $C_t$ & \bf Original & \bf Updated \\ \hline
$C_0$ & (food=eritrean) & (area=south, food=eritrean, pricerange=cheap) \\
$C_1$ & (area=south, food=eritrean) & (area=south, food=eritrean, pricerange=cheap)\\
$C_2$ & (area=south, food=spanish) & (area=south, food=spanish, pricerange=cheap) \\
$C_3$ & (area=south, food=spanish, pricerange=cheap) & (area=south, food=spanish, pricerange=cheap) \\
\hline
\end{tabular}
\end{center}
\caption{\label{tab:Goal Updates} An example of how DSTC2's turn-specific constraint labels can be transformed such that their behaviour can be replicated when training a dialogue manager.}
\end{table*}
The Goal Generator generates a random goal $G_0~\!=~\!(C_0,R)$ at the start of the dialogue. It consists of a set of \textit{constraints}, $C_0$, which specify the required venue e.g. (\texttt{food}=Spanish, \texttt{area}=north) and a number of \textit{requests}, $R$, that specify the information that the NUS wants about the final venue e.g. the address or the phone number. The possibilities for $C_t$ and $R$ are defined by the \textit{ontology}. In DSTC2 $C_t$ can consist of a maximum of three constraints; \texttt{food}, \texttt{area} and \texttt{pricerange}. Whether each of the three is present is independently sampled with a probability of 0.66, 0.62 and 0.58 respectively. These probabilities were estimated from the DSTC2 data set. If no constraint is sampled then the goal is re-sampled. For each slot in $C_0$ a value (e.g. north for \texttt{area}) is sampled uniformly from the ontology. Similarly, the presence of a request is independently sampled, followed by re-sampling if zero requests were chosen.

When training the sequence-to-sequence model, the Goal Generator is not used, but instead the goal labels from the DSTC2 dataset are used. In DSTC2 one goal-label is given to the entire dialogue. This goal is always the \textit{final} goal. If the user's goal at the start of the dialogue is (\texttt{food}=eritrean, \texttt{area}=south), which is changed to (\texttt{food}=spanish, \texttt{area}=south), due to the non-existence of an Eritrean restaurant in the south, using only the final goal is \textit{insufficient} to model the dialogue. The final goal can only be used for the requests as they are not altered during a dialogue. DSTC2 also provides turn-specific labels. These contain the constraints and requests expressed by the user up until and including the current turn. When training a policy with the NUS, such labels would not be available as they ``predict the future", i.e. when the turn-specific constraints change from (\texttt{area}=south) to (\texttt{food}=eritrean, \texttt{area}=south) it means that the user will inform the system about her desire to eat Eritrean food in the current turn. 

In related work on user-simulation for which the DSTC2 dataset was used, the final goal was used for the entire dialogue \cite{el2016sequence,Serras2017regularized,DBLP:conf/asru/LiuL17}. As stated above, we do not believe this to be sufficient. The following describes how to update the turn-specific constraint labels such that their behaviour can be replicated when training a DM's policy, whilst allowing goal changes to be modelled. The update strategy is illustrated in Table \ref{tab:Goal Updates} with an example. The final turn keeps its constraints, from which we iterate \textit{backwards} through the list of DSTC2's 
turn-specific constraints. The constraints of a turn will be set to the \textit{updated} constraints of the succeeding turn, besides if the same slot is present with a \textit{different} value. In that case the value will be kept. The behaviour of the updated turn-specific goal-labels can be replicated when the NUS is used to train a DM's policy. In the example, the food type changed due to the SDS expressing that there is no restaurant serving Eritrean food in the south. When deploying the NUS to train a policy, the goal is updated when the SDS outputs the \texttt{canthelp} dialogue act.
\subsection{Feature Extractor}\label{sec:FeaGen}
The Feature Extractor generates the feature vector that is appended to the sequence of feature vectors, here called \textit{Feature History}, that is passed to the sequence-to-sequence model. The input to the Feature Extractor is the output of the DM and the current goal $G_t$. Furthermore, as indicated in Figure \ref{fig:NUSfull}, the Feature Extractor keeps track of the currently accepted venue as well as the current and initial \textit{request-vector}, which is explained below. 

The feature vector $\mathbf{v}_t = [\mathbf{a}_t\: \mathbf{r}_t\: \mathbf{i}_t\: \mathbf{c}_t]$ is made up of four sub-vectors. The motivation behind the way in which these four vectors were designed is to provide an embedding for the system response that preserves all necessary \textit{value-independent} information.

The first vector, \textit{machine-act vector} $\mathbf{a}_t$, encodes the dialogue acts of the system response and consists of two parts; $\mathbf{a}_t \!=\! \left[\mathbf{a}^1_t\:\mathbf{a}^2_t\right]$. $\mathbf{a}^1_t$ is a binary representation of the system dialogue acts present in the input. Its length is thus the number of possible system dialogue acts. It is binary and not one-hot since in DSTC2 multiple dialogue acts can be in the system's response. $\mathbf{a}^2_t$ is a binary representation of the slot if the dialogue act is \texttt{request} or \texttt{select} and if it is \texttt{inform} or \texttt{expl-conf} together with a \textit{correct} slot-value pair for an informable slot. The length is four times the number of informable slots. $\mathbf{a}^2_t$ is necessary due to the dependence of the sentence structure on the exact slot mentioned by the system. The utterances of a user in response to \texttt{request(food)} and \texttt{request(area)} are often very different.

The second vector, \textit{request-vector} $\mathbf{r}_t$, is a binary representation of the requests that have not yet been fulfilled. It's length is thus the number of requestable slots. In comparison to the other three vectors the feature extractor needs to remember it for the next turn. At the start of the dialogue the indices corresponding to requests that are in $R$ are set to $1$ and the rest to $0$. Whenever the system informs 
a certain request the corresponding index in $\mathbf{r}_t$ is set to $0$. When a new venue is proposed $\mathbf{r}_t$ is reset to the original request vector, which is why the Feature Extractor keeps track of it.

The third vector, \textit{inconsistency-vector} $\mathbf{i}_t$, represents the inconsistency between the system's response and $C_t$. Every time a slot is mentioned by the system, when describing a venue (\texttt{inform}) or confirming a slot-value pair (\texttt{expl-conf} or \texttt{impl-conf}), the indices corresponding to the slots that have been misunderstood are set to 1. The length of $\mathbf{i}_t$ is the number of informable slots. This vector is necessary in order for the NUS to correct the system.

The fourth vector, $\mathbf{c}_t$, is a binary representation of the slots that are in the constraints $C_t$. It's length is thus the number of informable slots. This vector is necessary in order for the NUS to be able to inform about its preferred venue.

\subsection{Sequence-To-Sequence Model}
\begin{figure*}[htb!]
  \centering
  \centerline{\includegraphics[width = 0.95\linewidth]{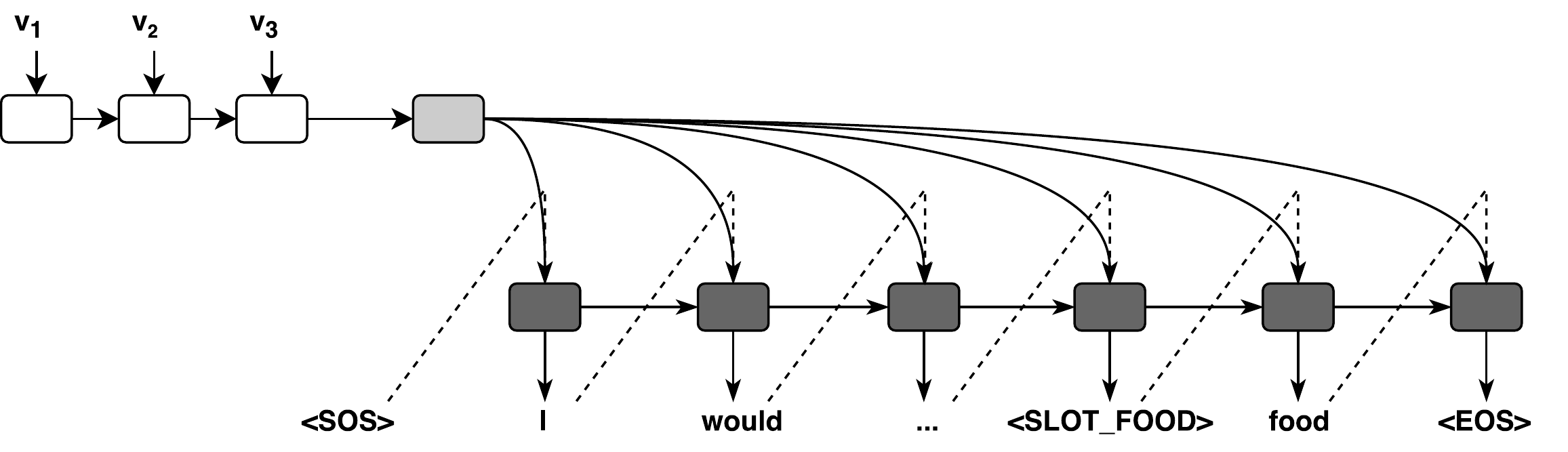}}
  \caption{Sequence-To-Sequence model of the Neural User Simulator. Here, the NUS is generating the user response to the third system output. The white, light-grey and dark blocks represent the RNN encoder, a fully-connected layer and the RNN decoder respectively. The previous output of the decoder is passed to its input for the next time-step. $\mathbf{v}_{3:1}$ are the first three feature vectors (see Sec. \ref{sec:FeaGen}).}
\label{fig:NUSseq2seq}
\end{figure*}
The sequence-to-sequence model (Figure \ref{fig:NUSseq2seq}) consists of an RNN encoder, followed by a fully-connect layer and an RNN decoder. An RNN can be defined as:
\begin{align}
\left(\mathbf{h}_t, \mathbf{s}_t\right) &= \textnormal{RNN}\left(\mathbf{x}_t, \mathbf{s}_{t-1}\right)
\end{align}
At time-step $t$, an RNN uses an input $\mathbf{x}_t$ and an internal state $\mathbf{s}_{t-1}$ to produce its output $\mathbf{h}_t$ and its new internal state $ \mathbf{s}_t$. A specific RNN-design is usually defined using matrix multiplications, element-wise additions and multiplications as well as element-wise non-linear functions. There are a plethora of different RNN architectures that could be used and explored. Given that such exploration is not the focus of this work a single layer LSTM \cite{hochreiter1997long} is used for both the RNN encoder and decoder. The exact LSTM version used in this work uses a forget gate without bias and does not use peep-holes.

The first RNN (shown as white blocks in Fig.~\ref{fig:NUSseq2seq}) takes one feature vector $\mathbf{v}_t$ at a time as its input ($\mathbf{x}^{E}_t = \mathbf{v}_t$). If the current dialogue turn is turn $T$ then the final output of the RNN encoder is given by $\mathbf{h}^{E}_{T}$, which is passed through a fully-connected layer (shown as the light-grey block) with linear activation function:
\begin{align}
\mathbf{p}_T &= W_{p}\mathbf{h}^{E}_{T} + \mathbf{b}_p
\end{align}
For a certain encoding $\mathbf{p}_T$ the sequence-to-sequence model should define a probability distribution over different sequences. By sampling from this distribution the NUS can generate a diverse set of sentences corresponding to the same dialogue context. The conditional probability distribution of a length $L$ sequence is defined as:
\begin{align}
\hspace{-0.2em} P\!\left( \mathbf{u} \!\mid\!  \mathbf{p} \right)\!&=\!P\!\left(w_0 \!\mid\! \mathbf{p}\right)\! \prod_{t=1}^{L}\!P\!\left(w_t \!\mid\! w_{t-1} ... w_0 , \mathbf{p} \right) \label{eq:seqProb}
\end{align}
The decoder RNN (shown as dark blocks) will be used to model $P\!\left(w_t \!\mid\! w_{t-1} ... w_0,\mathbf{p} \right)$. It's input at each time-step is the concatenation of an embedding $\mathbf{w}_{t-1}$ (we used 1-hot) of the previous word $w_{t-1}$ ($\mathbf{x}^{D}_t \!=\! [\mathbf{w}_{t-1} \: \mathbf{p}]$). For $P\!\left(w_0 \!\mid\! \mathbf{p}\right)$ a \textit{start-of-sentence} (\texttt{<SOS>}) token is used as $w_{-1}$. The end of the utterance is modelled using an \textit{end-of-sentence} (\texttt{<EOS>}) token. When the decoder RNN generates the \textit{end-of-sentence} token, the decoding process is terminated. The output of the decoder RNN, $\mathbf{h}^{D}_t$, is passed through an affine transform followed by the softmax function, \textit{SM}, to form $P\!\left(w_t \!\mid\! w_{t-1} ... w_0,\mathbf{p} \right)$. A word $w_t$ can be obtained by either taking the word with the highest probability or sampling from the distribution:
\begin{align}
P(w_{t} \mid w_{t-1} ... w_0,&\mathbf{p}) = SM(W_{w} \mathbf{h}^{D}_t + \mathbf{b}_w)\\
w_{t} \sim P(w_{t} &\mid w_{t-1} ... w_0,\mathbf{p})
\end{align}
During training the words are not sampled from the output distribution, but instead the true words from the dataset are used. This a common technique that is often referred to as \textit{teacher-forcing}, though it also directly follows from equation \ref{eq:seqProb}.

To generate a sequence using an RNN, beam-search is often used. Using beam-search with $n$ beams, the words corresponding to the top $n$ probabilities of $P\!\left(w_0 \!\mid\! \mathbf{p}\right)$ are the first $n$ beams. For each succeeding $w_t$, the $n$ words corresponding to the top $n$ probabilities of $P\!\left(w_{t} \!\mid\! w_{t-1} ... w_0,\mathbf{p}\right)$ are taken for each of the $n$ beams. This is followed by reducing the number of beams from now $n^2$ down to $n$, by taking the $n$ beams with the highest probability $P\!\left(w_{t} w_{t-1} ... w_0 \!\mid\! \mathbf{p}\right)$. This is a deterministic process. However, for the NUS to always give the same response in the same context is not realistic. Thus, the NUS cannot cover the full breadth of user behaviour if beam-search is used. 
To solve this issue while keeping the benefit of rejecting sequences with low probability, a type of beam-search with sampling is used.
The process is identical to the above, but $n$ words per beam are sampled from the probability distribution. The NUS is now non-deterministic resulting in a diverse US. Using 2 beams gave a good trade-off between reasonable responses and diversity.

\subsection{Training}
The neural sequence-to-sequence model is trained to maximize the log probability that it assigns to the user utterances of the training data set:
\begin{align}
\hspace{-0.2em}\mathcal{L}\!&=\!\sum_{n=1}^{N}\! \log\! P\!\left(w_0 \!\mid\! \mathbf{p}\right)\! \sum_{t=1}^{L_n} \log\! P\!\left(w_t \!\mid\! w_{t-1:0}, \mathbf{p} \right)
\end{align}
The network was implemented in Tensorflow \cite{tensorflow2015-whitepaper} and optimized using Tensorflow's default setup of the Adam optimizer \cite{kingma2014adam}. The LSTM layers and the fully-connected layer had widths of 100 each to give a reasonable number of overall parameters. The width was not tuned. The learning rate was optimised on a held out validation set and no regularization methods used. The training set was shuffled at the dialogue turn level.

The manual transcriptions of the DSTC2 training set (not the ASR output) were used to train the sequence-to-sequence model.
Since the transcriptions were done manually they contained spelling errors. These were manually corrected to ensure proper delexicalization. Some dialogues were discarded due to transcriptions errors being too large. After cleaning the dataset the training set consisted of 1609 dialogues with a total of 11638 dialogue turns. The validation set had 505 dialogues with 3896 dialogue turns. The maximum sequence length of the delexicalized turns was 22, including the end of sentence character. The maximum dialogue length was 30 turns.

\section{Experimental Setup}
The evaluation of 
user simulators is an ongoing area of research and a variety of techniques can be found in the literature. Most papers published on user simulation evaluate their US using \textit{direct} methods. These methods evaluate the US through a statistical measure of similarity between the outputs of the US and a real user on a test set. Multiple models can outperform the ABUS on these metrics. However, this is unsurprising since these user simulators were trained on the same or similar metrics. The ABUS was explicitly proposed as a tool to train the policy of a dialogue manager and it is still the dominant form of US used for this task. Therefore, the only fair comparison between a new US model and the ABUS is to use the \textit{indirect} method of evaluating the policies that were obtained by training with each US.

\subsection{Training}
All dialogue policies were trained with the PyDial toolkit \cite{ultes2017pydial}, by interacting with either the NUS or 
ABUS. The RL algorithm used is GP-SARSA \cite{gasic2014gaussian} with hyperparameters taken from \cite{Casanueva2017benchenv}. 
The reward function used gives a reward of 20 to a successfully completed dialogue and of -1 for each dialogue turn. The maximum dialogue length was 25 turns. The presented metrics are success rate (SR) and average reward over test dialogues. SR is the percentage of dialogues for which the system satisfied both the user's constraints and requests. The final goal, after possible goal changes, was used for this evaluation. When policies are trained using the NUS, its output is parsed using PyDial's regular expression based semantic decoder. The policies were trained for 4000 dialogues.

\subsection{Testing with a simulated user}
In Schatzmann et.~al \shortcite{schatztmann2005effects} \textit{cross-model evaluation} is proposed to compare user simulators. First, the user simulators to be evaluated are used to train $N$ policy each. Then these policies are tested using the different user simulators and the results averaged. \citet{schatztmann2005effects} showed that a strategy learned with a good user model still performs well when tested on poor user models. If a policy performs well on all user simulators and not just on the one that it was trained on, it indicates that the US with which it was trained is diverse and realistic, and thus the policy is likely to perform better on real users. For each US five policies ($N=5$), each using a different random seed for initialisation, are trained. Results are reported for both the best and the average performance on 1000 test dialogues. The ABUS is programmed to always mention the new goal after a goal change. In order to not let this affect our results we implement the same for the NUS by re-sampling a sentence if the new goal is not mentioned.

\subsection{Testing with real users}
Though the above test is already more indicative of policy performance on real users than measuring statistical metrics of user behaviour, a better test is to 
test with human users. For the test on human users, two policies for each US that was used for training are chosen from the five policies. The first policy is the one that performed best when \textit{tested on the NUS}. The second is the one that performed best when \textit{tested on the ABUS}. This choice of policies is motivated by a type of overfitting to be seen  in Sec.~\ref{sec:crossmodeleval}. The evaluation of the trained dialogue policies in interaction with real users follows a similar set-up to \cite{jurvcivcek2011real}. Users are recruited through the Amazon Mechanical Turk (AMT) service. 1000 dialogues (250 per policy) were gathered. The learnt policies were incorporated into an SDS pipeline with a commercial ASR system. The AMT users were asked to find a restaurant that matches certain constraints and find certain requests. Subjects were randomly allocated to one of the four analysed systems. After each dialogue the users were asked whether they judged the dialogue to be successful or not which was then translated to the reward measure.

\section{Experimental Results}
\subsection{Cross-Model Evaluation}\label{sec:crossmodeleval}
\begin{table}
\begin{tabular}{l|ll|ll}
Train. Sim. & \multicolumn{4}{c}{Eval. Sim.}\\
   &  \multicolumn{2}{c}{NUS} & \multicolumn{2}{c}{ABUS} \\
 		& Rew. & Suc. & Rew. & Suc. \\ \hline
NUS-best 		& 13.0 & 98.0\textsuperscript{$\mathcal{N}_1$} & 13.3 & 99.8 \\ \hline
ABUS-best		& 1.53 & 71.5\textsuperscript{$\mathcal{A}_1$} & 13.8 & 99.9\textsuperscript{$\mathcal{A}_2$} \\ \hline
NUS-avg 	& 12.4 & 96.6 & 11.2 & 94.0 \\ \hline
ABUS-avg	& -7.6 & 45.5 & 13.5 & 99.5  \\ \hline
\end{tabular}
\caption{Results for policies trained for 4000 dialogues on NUS and ABUS when tested on both USs for 1000 dialogues. Five policies with different initialisations were trained for each US. Both average and best results are shown.
}
\label{tab:crossmodel20}
\end{table}
Table~\ref{tab:crossmodel20} shows the results of the cross-model evaluation after 4000 training dialogues. The policies trained with the NUS achieved an average success rate (SR) of 94.0\% and of 96.6\% when tested on the ABUS and the NUS, respectively. By comparison, the policies trained with the ABUS achieved average SRs of 99.5\% and 45.5\% respectively. Thus, training with the NUS leads to policies that can perform well on both USs, which is not the case for training with the ABUS. Furthermore, the best SRs when tested on the ABUS are similar at 99.9\% (ABUS) and 99.8\% (NUS). When tested on the NUS the best SRs were 71.5\% (ABUS) and 98.0\% (NUS). This shows that the behaviour of the Neural User Simulator is realistic and diverse enough to train policies that can also perform very well on the Agenda-Based User Simulator.

Of the five policies, for each US, the policy performing best on the NUS was not the best performing policy on the ABUS. This could indicate that the policy ``overfits'' to a particular user simulator. Overfitting usually manifests itself in worse results as the model is trained for longer.
Five policies trained on each US for only 1000 dialogues were also evaluated, the results of which can be seen in Table~\ref{tab:crossmodel5}. After training for 1000 dialogues, the average SR of the policies trained on the NUS when tested on the ABUS was 97.3\% in comparison to 94.0\% after 4000 dialogues. This behaviour was observed for all five seeds, which indicates that the policy indeed overfits to the NUS. For the policies trained with the ABUS this was not observed. This could indicate that the policy can learn to exploit some of the shortcomings of the trained NUS.

\begin{table}
\begin{tabular}{l|lr|ll}
Train. Sim.& \multicolumn{4}{c}{Eval. Sim.}\\
   &  \multicolumn{2}{c}{NUS} & \multicolumn{2}{c}{ABUS} \\
 		& Rew.& Suc.& Rew.&Suc.\\ \hline
NUS-best 		& 12.2 & 95.9 & 13.9 & 99.9\textsuperscript{$\mathcal{N}_2$} \\\hline
ABUS-best		& -4.0 & 54.8 & 13.2 & 99.0 \\\hline
NUS-avg 	& 12.0 & 95.4 & 12.2 & 97.3 \\\hline
ABUS-avg	& -9.48 & 42.3 & 12.8 & 98.4 \\\hline
\end{tabular}
\caption{As Table \ref{tab:crossmodel20} but 
trained for 1000 dialogues.}
\label{tab:crossmodel5}
\end{table}

\subsection{Human Evaluation}
The results of the human evaluation are shown in Table~\ref{tab:realusers} for 250 dialogues per policy. In Table~\ref{tab:realusers} policies are marked using an ID ($\mathcal{U}_{\alpha}$) that translates to results 
in Tables~\ref{tab:crossmodel20} and \ref{tab:crossmodel5}. Both policies trained with the NUS outperformed those trained on the ABUS in terms of both reward and success rate. The best performing policy trained on the NUS achieves a 93.4\% success rate and 13.8 average rewards whilst the best performing policy trained with the ABUS achieves only a 90.0\% success rate and 13.3 average reward. This shows that the good performance of the NUS on the cross-model evaluation transfers to real users. Furthermore, the overfitting to a particular US is also observed in the real user evaluation. For not only the policies trained on the NUS, but also those trained on the ABUS, the best performing policy was the policy that performed best on the other US.
\begin{table}
\begin{tabular}{l|lr}
Training Simulator   &  \multicolumn{2}{c}{Human Evaluation}  \\
 		& Rew. & Suc. \\ \hline
NUS - $\mathcal{N}_1$  & 13.4 & 91.8 \\\hline
NUS - $\mathcal{N}_2$ & 13.8 & 93.4 \\\hline
ABUS - $\mathcal{A}_1$  & 13.3 & 90.0 \\\hline
ABUS - $\mathcal{A}_2$ & 13.1 & 88.5 \\\hline
\end{tabular}
\caption{Real User Evaluation. Results over 250 dialogues with human users. $\mathcal{N}_1$ and $\mathcal{A}_1$ performed best on the NUS. $\mathcal{N}_2$ and $\mathcal{A}_2$ performed best on the ABUS. Rewards are not comparable to Table \ref{tab:crossmodel20} and \ref{tab:crossmodel5} since all user goals were achievable.}
\label{tab:realusers}
\end{table}

\section{Conclusion}
We introduced the Neural User Simulator (NUS), which uses the system's response in its semantic form as input and gives a natural language response. It thus needs \textit{less labelling} of the training data than User Simulators that generate a response in semantic form. It was shown that the NUS learns realistic user behaviour from a corpus of recorded dialogues such that it can be used to optimise the policy of the dialogue manager of a spoken dialogue system. The NUS was compared to the Agenda-Based User Simulator by evaluating policies trained with these user simulators. The trained policies were compared both by testing them with simulated users and also with real users. The NUS excelled on both evaluation tasks.

\section*{Acknowledgements}
This research was partly funded by the EPSRC grant EP/M018946/1 Open Domain Statistical Spoken Dialogue Systems. Florian Kreyssig is supported by the Studienstiftung des Deutschen Volkes. Pawe{\l} Budzianowski is supported by the EPSRC and Toshiba Research Europe Ltd.

\bibliography{acl2018}

\begin{thebibliography}{41}
\expandafter\ifx\csname natexlab\endcsname\relax\def\natexlab#1{#1}\fi

\bibitem[{Abadi et~al.(2015)Abadi, Agarwal, Barham, Brevdo, Chen, Citro,
  Corrado, Davis, Dean, Devin, Ghemawat, Goodfellow, Harp, Irving, Isard, Jia,
  Jozefowicz, Kaiser, Kudlur, Levenberg, Man\'{e}, Monga, Moore, Murray, Olah,
  Schuster, Shlens, Steiner, Sutskever, Talwar, Tucker, Vanhoucke, Vasudevan,
  Vi\'{e}gas, Vinyals, Warden, Wattenberg, Wicke, Yu, and
  Zheng}]{tensorflow2015-whitepaper}
Mart\'{\i}n Abadi, Ashish Agarwal, Paul Barham, Eugene Brevdo, Zhifeng Chen,
  Craig Citro, Greg~S. Corrado, Andy Davis, Jeffrey Dean, Matthieu Devin,
  Sanjay Ghemawat, Ian Goodfellow, Andrew Harp, Geoffrey Irving, Michael Isard,
  Yangqing Jia, Rafal Jozefowicz, Lukasz Kaiser, Manjunath Kudlur, Josh
  Levenberg, Dandelion Man\'{e}, Rajat Monga, Sherry Moore, Derek Murray, Chris
  Olah, Mike Schuster, Jonathon Shlens, Benoit Steiner, Ilya Sutskever, Kunal
  Talwar, Paul Tucker, Vincent Vanhoucke, Vijay Vasudevan, Fernanda Vi\'{e}gas,
  Oriol Vinyals, Pete Warden, Martin Wattenberg, Martin Wicke, Yuan Yu, and
  Xiaoqiang Zheng. 2015.
\newblock \href {https://www.tensorflow.org/} {{TensorFlow}: Large-scale
  machine learning on heterogeneous systems}.
\newblock Software available from tensorflow.org.

\bibitem[{Casanueva et~al.(2017)Casanueva, Budzianowski, Su, Mrk{\v{s}}i{\'c},
  Wen, Ultes, Rojas-Barahona, Young, and
  Ga{\v{s}}i{\'c}}]{Casanueva2017benchenv}
I{\~n}igo Casanueva, Pawe{\l} Budzianowski, Pei-Hao Su, Nikola
  Mrk{\v{s}}i{\'c}, Tsung-Hsien Wen, Stefan Ultes, Lina Rojas-Barahona, Steve
  Young, and Milica Ga{\v{s}}i{\'c}. 2017.
\newblock A benchmarking environment for reinforcement learning based task
  oriented dialogue management.
\newblock In \emph{NIPS Deep Reinforcement Learning Symposium}.

\bibitem[{Casanueva et~al.(2018)Casanueva, Budzianowski, Su, Ultes,
  Rojas-Barahona, Tseng, and Ga{\v{s}}i{\'c}}]{Casanueva2018feudal}
I{\~n}igo Casanueva, Pawe{\l} Budzianowski, Pei-Hao Su, Stefan Ultes, Lina
  Rojas-Barahona, Bo-Hsiang Tseng, and Milica Ga{\v{s}}i{\'c}. 2018.
\newblock Feudal reinforcement learning for dialogue management in large
  domains.
\newblock In \emph{Proc. NAACL 2018}.

\bibitem[{Chandramohan et~al.(2011)Chandramohan, Geist, Lefevre, and
  Pietquin}]{chandramohan2011user}
Senthilkumar Chandramohan, Matthieu Geist, Fabrice Lefevre, and Olivier
  Pietquin. 2011.
\newblock User simulation in dialogue systems using inverse reinforcement
  learning.
\newblock In \emph{Proceedings of the Twelfth Annual Conference of the
  International Speech Communication Association}.

\bibitem[{Chang et~al.(2017)Chang, Yang, Chen, Zhou, and
  Yu}]{chang2017affordable}
Cheng Chang, Runzhe Yang, Lu~Chen, Xiang Zhou, and Kai Yu. 2017.
\newblock \href {http://aclweb.org/anthology/D17-1234} {Affordable on-line
  dialogue policy learning}.
\newblock In \emph{Proceedings of the 2017 Conference on Empirical Methods in
  Natural Language Processing}, pages 2200--2209.

\bibitem[{Chen et~al.(2017)Chen, Zhou, Chang, Yang, and Yu}]{chen2017agent}
Lu~Chen, Xiang Zhou, Cheng Chang, Runzhe Yang, and Kai Yu. 2017.
\newblock \href {http://aclweb.org/anthology/D17-1260} {Agent-aware dropout dqn
  for safe and efficient on-line dialogue policy learning}.
\newblock In \emph{Proceedings of the 2017 Conference on Empirical Methods in
  Natural Language Processing}, pages 2454--2464.

\bibitem[{Crook and Marin(2017)}]{crook2017sequence}
Paul Crook and Alex Marin. 2017.
\newblock Sequence to sequence modeling for user simulation in dialog systems.
\newblock In \emph{Proceedings of the 18th Annual Conference of the
  International Speech Communication Association}.

\bibitem[{Cuay{\'a}huitl et~al.(2005)Cuay{\'a}huitl, Renals, Lemon, and
  Shimodaira}]{cuayahuitl2005human}
Heriberto Cuay{\'a}huitl, Steve Renals, Oliver Lemon, and Hiroshi Shimodaira.
  2005.
\newblock Human-computer dialogue simulation using hidden markov models.
\newblock In \emph{Automatic Speech Recognition and Understanding, 2005 IEEE
  Workshop on}, pages 290--295. IEEE.

\bibitem[{Eckert et~al.(1997)Eckert, Levin, and Pieraccini}]{eckert1997user}
Wieland Eckert, Esther Levin, and Roberto Pieraccini. 1997.
\newblock User modeling for spoken dialogue system evaluation.
\newblock In \emph{Automatic Speech Recognition and Understanding, 1997.
  Proceedings., 1997 IEEE Workshop on}, pages 80--87. IEEE.

\bibitem[{El~Asri et~al.(2016)El~Asri, He, and Suleman}]{el2016sequence}
Layla El~Asri, Jing He, and Kaheer Suleman. 2016.
\newblock A sequence-to-sequence model for user simulation in spoken dialogue
  systems.
\newblock \emph{Proceedings of the 17th Annual Conference of the International
  Speech Communication Association}, pages 1151--1155.

\bibitem[{Fatemi et~al.(2016)Fatemi, El~Asri, Schulz, He, and
  Suleman}]{fatemi2016policy}
Mehdi Fatemi, Layla El~Asri, Hannes Schulz, Jing He, and Kaheer Suleman. 2016.
\newblock \href {http://www.aclweb.org/anthology/W16-3613} {Policy networks
  with two-stage training for dialogue systems}.
\newblock In \emph{Proceedings of the 17th Annual Meeting of the Special
  Interest Group on Discourse and Dialogue}, pages 101--110.

\bibitem[{Ga{\v{s}}i{\'c} and Young(2014)}]{gasic2014gaussian}
Milica Ga{\v{s}}i{\'c} and Steve Young. 2014.
\newblock Gaussian processes for pomdp-based dialogue manager optimization.
\newblock \emph{IEEE/ACM Transactions on Audio, Speech, and Language
  Processing}, 22(1):28--40.

\bibitem[{Ga\v{s}i\'{c} et~al.(2011)Ga\v{s}i\'{c}, Jur\v{c}\'{i}\v{c}ek,
  Thomson, Yu, and Young}]{gjtk11}
M.~Ga\v{s}i\'{c}, F.~Jur\v{c}\'{i}\v{c}ek, B.~Thomson, K.~Yu, and S.~Young.
  2011.
\newblock On-line policy optimisation of spoken dialogue systems via live
  interaction with human subjects.
\newblock In \emph{Automatic Speech Recognition and Understanding, 2011 IEEE
  Workshop on}.

\bibitem[{Georgila et~al.(2005)Georgila, Henderson, and
  Lemon}]{georgila2005learning}
Kallirroi Georgila, James Henderson, and Oliver Lemon. 2005.
\newblock Learning user simulations for information state update dialogue
  systems.
\newblock In \emph{Ninth European Conference on Speech Communication and
  Technology}.

\bibitem[{Goldwater et~al.(2010)Goldwater, Jurafsky, and
  Manning}]{goldwater2010words}
Sharon Goldwater, Dan Jurafsky, and Christopher~D Manning. 2010.
\newblock Which words are hard to recognize? prosodic, lexical, and disfluency
  factors that increase speech recognition error rates.
\newblock \emph{Speech Communication}, 52(3):181--200.

\bibitem[{Henderson et~al.(2014)Henderson, Thomson, and
  Williams}]{henderson2014second}
Matthew Henderson, Blaise Thomson, and Jason~D Williams. 2014.
\newblock \href {http://www.aclweb.org/anthology/W14-4337} {The second dialog
  state tracking challenge}.
\newblock In \emph{Proceedings of the 15th Annual Meeting of the Special
  Interest Group on Discourse and Dialogue (SIGDIAL)}, pages 263--272.

\bibitem[{Hochreiter and Schmidhuber(1997)}]{hochreiter1997long}
Sepp Hochreiter and J{\"u}rgen Schmidhuber. 1997.
\newblock Long short-term memory.
\newblock \emph{Neural computation}, 9(8):1735--1780.

\bibitem[{Jur{\v{c}}{\'\i}{\v{c}}ek et~al.(2011)Jur{\v{c}}{\'\i}{\v{c}}ek,
  Keizer, Ga{\v{s}}i{\'c}, Mairesse, Thomson, Yu, and
  Young}]{jurvcivcek2011real}
Filip Jur{\v{c}}{\'\i}{\v{c}}ek, Simon Keizer, Milica Ga{\v{s}}i{\'c}, Francois
  Mairesse, Blaise Thomson, Kai Yu, and Steve Young. 2011.
\newblock Real user evaluation of spoken dialogue systems using amazon
  mechanical turk.
\newblock In \emph{Proceedings of the Twelfth Annual Conference of the
  International Speech Communication Association}.

\bibitem[{Kingma and Ba(2015)}]{kingma2014adam}
Diederik~P Kingma and Jimmy Ba. 2015.
\newblock Adam: A method for stochastic optimization.
\newblock In \emph{Proc. ICLR}, San Diego.

\bibitem[{Koo et~al.(2015)Koo, Ryu, and Lee}]{koo2015implementation}
Sangjun Koo, Seonghan Ryu, and Gary~Geunbae Lee. 2015.
\newblock Implementation of generic positive-negative tracker in extensible
  dialog system.
\newblock In \emph{Automatic Speech Recognition and Understanding (ASRU), 2015
  IEEE Workshop on}, pages 798--805. IEEE.

\bibitem[{Li et~al.(2017)Li, Chen, Li, Gao, and Celikyilmaz}]{li2017end}
Xiujun Li, Yun-Nung Chen, Lihong Li, Jianfeng Gao, and Asli Celikyilmaz. 2017.
\newblock \href {http://www.aclweb.org/anthology/I17-1074} {End-to-end
  task-completion neural dialogue systems}.
\newblock In \emph{Proceedings of the Eighth International Joint Conference on
  Natural Language Processing (Volume 1: Long Papers)}, pages 733--743, Taipei,
  Taiwan. Asian Federation of Natural Language Processing.

\bibitem[{Liu and Lane(2017)}]{DBLP:conf/asru/LiuL17}
Bing Liu and Ian Lane. 2017.
\newblock Iterative policy learning in end-to-end trainable task-oriented
  neural dialog models.
\newblock In \emph{2017 {IEEE} Automatic Speech Recognition and Understanding
  Workshop, {ASRU}}, pages 482--489.

\bibitem[{Mrk{\v{s}}i{\'c} et~al.(2017)Mrk{\v{s}}i{\'c}, S{\'e}aghdha, Wen,
  Thomson, and Young}]{mrkvsic2017neural}
Nikola Mrk{\v{s}}i{\'c}, Diarmuid~{\'O} S{\'e}aghdha, Tsung-Hsien Wen, Blaise
  Thomson, and Steve Young. 2017.
\newblock \href {http://www.aclweb.org/anthology/P17-1163} {Neural belief
  tracker: Data-driven dialogue state tracking}.
\newblock In \emph{Proceedings of the 55th Annual Meeting of the Association
  for Computational Linguistics (Volume 1: Long Papers)}, volume~1, pages
  1777--1788.

\bibitem[{Pietquin and Dutoit(2006)}]{pietquin2006probabilistic}
Olivier Pietquin and Thierry Dutoit. 2006.
\newblock A probabilistic framework for dialog simulation and optimal strategy
  learning.
\newblock \emph{IEEE Transactions on Audio, Speech, and Language Processing},
  14(2):589--599.

\bibitem[{Ramadan et~al.(2018)Ramadan, Budzianowski, and
  Ga{\v{s}}i{\'c}}]{ramadan2018large}
Osman Ramadan, Pawe{\l} Budzianowski, and Milica Ga{\v{s}}i{\'c}. 2018.
\newblock Large-scale multi-domain belief tracking with knowledge sharing.
\newblock In \emph{Proceedings of the 56th Annual Meeting of the Association
  for Computational Linguistics (Volume 2: Short Papers)}. Association for
  Computational Linguistics.

\bibitem[{Roy et~al.(2000)Roy, Pineau, and Thrun}]{roy2000spoken}
Nicholas Roy, Joelle Pineau, and Sebastian Thrun. 2000.
\newblock \href {http://aclweb.org/anthology/P/P00/P00-1013.pdf} {Spoken
  dialogue management using probabilistic reasoning}.
\newblock In \emph{Proceedings of the 38th Annual Meeting on Association for
  Computational Linguistics}, pages 93--100. Association for Computational
  Linguistics.

\bibitem[{Schatzmann et~al.(2007)Schatzmann, Thomson, Weilhammer, Ye, and
  Young}]{schatzmann2007agenda}
Jost Schatzmann, Blaise Thomson, Karl Weilhammer, Hui Ye, and Steve Young.
  2007.
\newblock \href {http://www.aclweb.org/anthology/N07-2038} {Agenda-based user
  simulation for bootstrapping a pomdp dialogue system}.
\newblock In \emph{Human Language Technologies 2007: The Conference of the
  North American Chapter of the Association for Computational Linguistics;
  Companion Volume, Short Papers}, pages 149--152. Association for
  Computational Linguistics.

\bibitem[{Schatztmann et~al.(2005)Schatztmann, Stuttle, Weilhammer, and
  Young}]{schatztmann2005effects}
Jost Schatztmann, Matthew~N Stuttle, Karl Weilhammer, and Steve Young. 2005.
\newblock Effects of the user model on simulation-based learning of dialogue
  strategies.
\newblock In \emph{Automatic Speech Recognition and Understanding, 2005 IEEE
  Workshop on}, pages 220--225. IEEE.

\bibitem[{Scheffler and Young(2000)}]{scheffler2000probabilistic}
Konrad Scheffler and Steve Young. 2000.
\newblock Probabilistic simulation of human-machine dialogues.
\newblock In \emph{Acoustics, Speech, and Signal Processing, 2000. ICASSP'00.
  Proceedings. 2000 IEEE International Conference on}, volume~2, pages
  II1217--II1220. IEEE.

\bibitem[{Scheffler and Young(2001)}]{scheffler2001corpus}
Konrad Scheffler and Steve Young. 2001.
\newblock Corpus-based dialogue simulation for automatic strategy learning and
  evaluation.
\newblock In \emph{Proc. NAACL Workshop on Adaptation in Dialogue Systems},
  pages 64--70.

\bibitem[{Serras et~al.(2017)Serras, Torres, and del
  Pozo}]{Serras2017regularized}
Manex Serras, Mar{\'i}a In{\'e}s~Torres Torres, and Arantza del Pozo. 2017.
\newblock Regularized neural user model for goal oriented spoken dialogue
  systems.
\newblock In \emph{International Workshop on Spoken Dialogue Systems}.
  Association for Computational Linguistics.

\bibitem[{Shah et~al.(2018)Shah, Hakkani-T{\"u}r, T{\"u}r, Rastogi, Bapna,
  Nayak, and Heck}]{shah2018building}
Pararth Shah, Dilek Hakkani-T{\"u}r, Gokhan T{\"u}r, Abhinav Rastogi, Ankur
  Bapna, Neha Nayak, and Larry Heck. 2018.
\newblock Building a conversational agent overnight with dialogue self-play.
\newblock \emph{arXiv preprint arXiv:1801.04871}.

\bibitem[{Sun et~al.(2014)Sun, Chen, Zhu, and Yu}]{sun2014sjtu}
Kai Sun, Lu~Chen, Su~Zhu, and Kai Yu. 2014.
\newblock \href {http://www.aclweb.org/anthology/W/W14/W14-4343.pdf} {The sjtu
  system for dialog state tracking challenge 2}.
\newblock In \emph{Proceedings of the 15th Annual Meeting of the Special
  Interest Group on Discourse and Dialogue (SIGDIAL)}, pages 318--326.

\bibitem[{Sun et~al.(2016)Sun, Xie, and Yu}]{sun2016recurrent}
Kai Sun, Qizhe Xie, and Kai Yu. 2016.
\newblock Recurrent polynomial network for dialogue state tracking.
\newblock \emph{Dialogue \& Discourse}, 7(3):65--88.

\bibitem[{Sutskever et~al.(2014)Sutskever, Vinyals, and
  Le}]{sutskever2014sequence}
Ilya Sutskever, Oriol Vinyals, and Quoc~V Le. 2014.
\newblock Sequence to sequence learning with neural networks.
\newblock In \emph{Advances in neural information processing systems}, pages
  3104--3112.

\bibitem[{Ultes et~al.(2017)Ultes, Rojas~Barahona, Su, Vandyke, Kim, Casanueva,
  Budzianowski, Mrk\v{s}i\'{c}, Wen, Ga{\v{s}}i{\'c}, and
  Young}]{ultes2017pydial}
Stefan Ultes, Lina~M. Rojas~Barahona, Pei-Hao Su, David Vandyke, Dongho Kim,
  I\~{n}igo Casanueva, Pawe{\l} Budzianowski, Nikola Mrk\v{s}i\'{c},
  Tsung-Hsien Wen, Milica Ga{\v{s}}i{\'c}, and Steve Young. 2017.
\newblock \href {http://aclweb.org/anthology/P17-4013} {{PyDial: A Multi-domain
  Statistical Dialogue System Toolkit}}.
\newblock In \emph{Proceedings of ACL 2017, System Demonstrations}, pages
  73--78, Vancouver, Canada. Association for Computational Linguistics.

\bibitem[{Weisz et~al.(2018)Weisz, Budzianowski, Su, and
  Ga{\v{s}}i{\'c}}]{weisz2018sample}
Gell{\'e}rt Weisz, Pawe{\l} Budzianowski, Pei-Hao Su, and Milica
  Ga{\v{s}}i{\'c}. 2018.
\newblock Sample efficient deep reinforcement learning for dialogue systems
  with large action spaces.
\newblock \emph{arXiv preprint arXiv:1802.03753}.

\bibitem[{Williams(2008)}]{williams2008evaluating}
Jason~D Williams. 2008.
\newblock Evaluating user simulations with the cram{\'e}r--von mises
  divergence.
\newblock \emph{Speech communication}, 50(10):829--846.

\bibitem[{Williams and Young(2007)}]{williams2007partially}
Jason~D Williams and Steve Young. 2007.
\newblock Partially observable markov decision processes for spoken dialog
  systems.
\newblock \emph{Computer Speech \& Language}, 21(2):393--422.

\bibitem[{Young et~al.(2013)Young, Ga{\v{s}}i{\'c}, Thomson, and
  Williams}]{young2013pomdp}
Steve Young, Milica Ga{\v{s}}i{\'c}, Blaise Thomson, and Jason~D Williams.
  2013.
\newblock Pomdp-based statistical spoken dialog systems: A review.
\newblock \emph{Proceedings of the IEEE}, 101(5):1160--1179.

\bibitem[{Zilka and Jurcicek(2015)}]{zilka2015incremental}
Lukas Zilka and Filip Jurcicek. 2015.
\newblock Incremental lstm-based dialog state tracker.
\newblock In \emph{Automatic Speech Recognition and Understanding (ASRU), 2015
  IEEE Workshop on}, pages 757--762. IEEE.

\end{thebibliography}
\bibliographystyle{acl_natbib}

\end{document}